\pdfoutput=1

\documentclass[11pt]{article}

\usepackage[final]{acl}
\usepackage{tabularx, url}
\usepackage{multicol,multirow}
\usepackage{booktabs, pifont}
\usepackage{times}
\usepackage{latexsym}
\usepackage[labelformat=empty]{subcaption}
\usepackage{amsmath, amssymb}
\usepackage{color, colortbl} 

\usepackage[T1]{fontenc}

\usepackage[utf8]{inputenc}

\usepackage{microtype}

\usepackage{inconsolata}

\usepackage{graphicx}

\title{Universal Acoustic Adversarial Attacks for Flexible Control of Speech-LLMs}

\author{Rao Ma, Mengjie Qian, Vyas Raina, Mark Gales, Kate Knill \\
  ALTA Institute, Department of Engineering, University of Cambridge \\
  \texttt{\{rm2114,mq227,vr313,mjfg100,kmk1001\}@cam.ac.uk} \\}

\definecolor{Gray}{gray}{0.9}

\begin{document}
\maketitle
\begin{abstract}
The combination of pre-trained speech encoders with large language models has enabled the development of speech LLMs that can handle a wide range of spoken language processing tasks. While these models are powerful and flexible, this very flexibility may make them more vulnerable to adversarial attacks. To examine the extent of this problem, in this work we investigate universal acoustic adversarial attacks on speech LLMs. Here a fixed, universal, adversarial audio segment is prepended to the original input audio. We initially investigate attacks that cause the model to either produce no output or to perform a modified task overriding the original prompt. We then extend the nature of the attack to be selective so that it activates only when specific input attributes, such as a speaker gender or spoken language, are present. Inputs without the targeted attribute should be unaffected, allowing fine-grained control over the model outputs. Our findings reveal critical vulnerabilities in Qwen2-Audio and Granite-Speech and suggest that similar speech LLMs may be susceptible to universal adversarial attacks. This highlights the need for more robust training strategies and improved resistance to adversarial attacks.
\end{abstract}

\section{Introduction}

Large language models (LLMs) have demonstrated remarkable language understanding and generation capabilities~\cite{touvron2023llama, achiam2023gpt, team2023gemini}. Recent methods have extended these language models to accommodate the speech modality by aligning pre-trained speech encoders with LLMs~\cite{fathullah2024prompting, yu2024connecting, tang2024salmonn, chu2024qwen2}. 
Unlike pre-trained speech models like Whisper \cite{radford2023robust}, which are limited to automatic speech recognition (ASR) and speech translation into English, speech LLMs offer greater flexibility by enabling instruction following across a wide range of tasks. This makes them more powerful and versatile, particularly in interactive and open-ended audio understanding and reasoning tasks.
However, the enhanced capabilities of speech LLMs may also expose them to adversarial threats that exploit their flexibility~\cite{hughesattacking, yang2024audio}, raising critical concerns about their robustness in real-world applications. Despite their increasing adoption, these vulnerabilities remain largely underexplored.

\begin{figure}[t]
    \centering
    \includegraphics[width=1.0\linewidth]{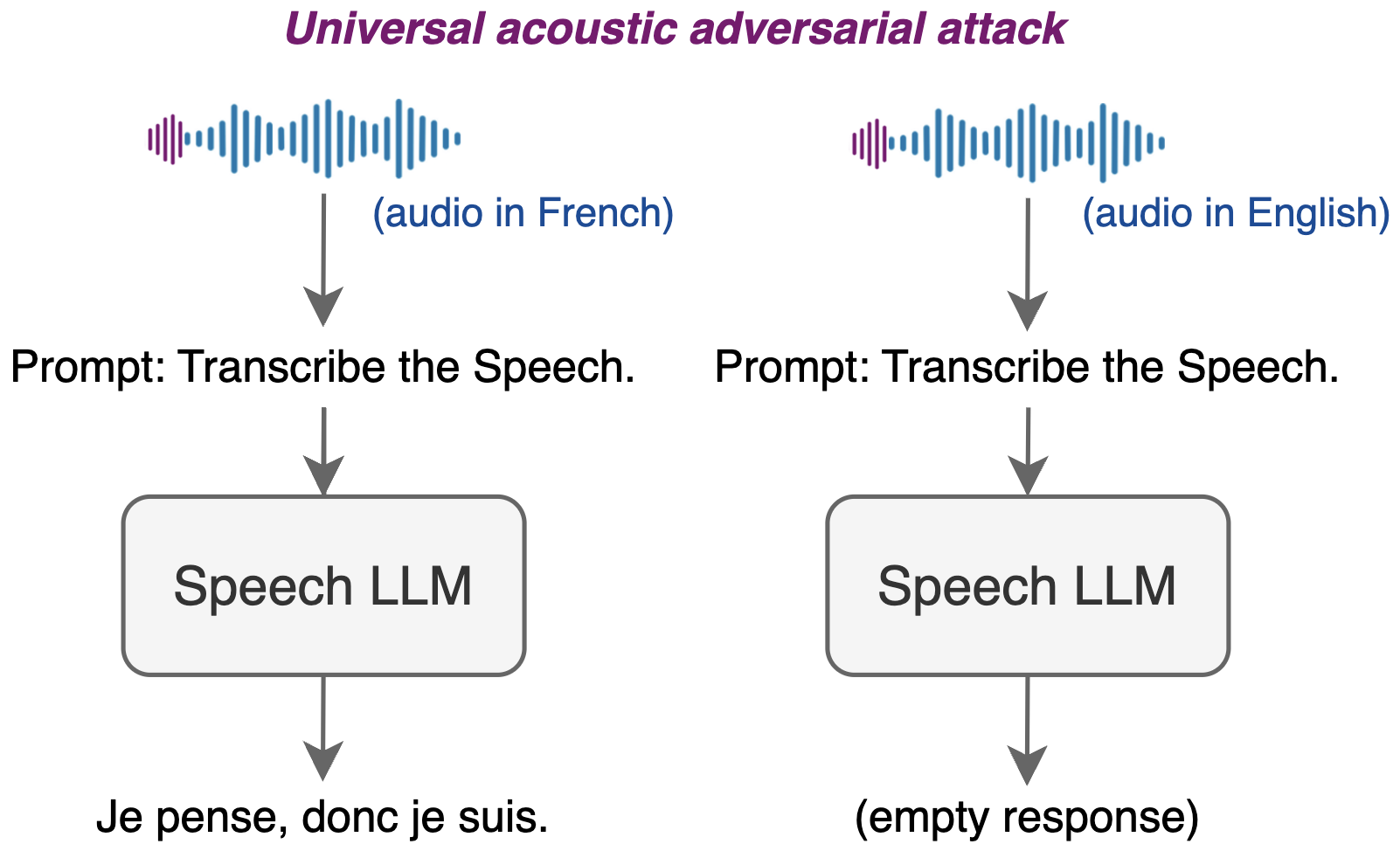}
    \caption{Illustration of language-based selective attack.}
\end{figure}

Previous work on Whisper~\cite{raina2024muting, raina2024controlling} proposed a novel universal acoustic adversarial attack, which learns a fixed audio segment capable of manipulating the model's behaviour. Specifically, these studies demonstrated that the attack can be designed to \textit{mute} the model so it outputs nothing, or it can manipulate the model to perform a different task entirely than instructed. The attack approach is both efficient and flexible, as it can universally be prepended to any input audio without altering its representation, and it transfers well across different settings. Building on this approach, we extend the investigation to powerful and representative speech LLMs, Qwen2-Audio~\cite{chu2024qwen2} and Granite-Speech~\cite{granite_speech_2025}, and conduct a comprehensive analysis of their vulnerabilities to universal acoustic attacks. 

In this work, we begin by evaluating the effectiveness of \textbf{general attacks} on speech LLMs, focusing on two primary forms: muting outputs and task control. These attacks involve applying the same fixed adversarial audio segment to all inputs, aiming to disrupt the model outputs. In the muting scenario, the adversarial audio causes the model to return empty responses. This could pose serious risks, particularly if the speech contains harmful or policy-violating content and the attack is used to evade platform moderation. In the task control setting, the attack does not silence the model but instead overrides the original prompt, causing the model to perform a different task than intended. This form of attack reveals a deeper vulnerability in the model’s internal task alignment mechanisms. By hijacking behaviour without disabling the model entirely, task control attacks are subtle, harder to detect, and potentially more damaging.

Building on these general attacks, we introduce a novel form of adversarial manipulation, which we refer to as a \textbf{selective attack}. Unlike universal attacks that affect all inputs indiscriminately, selective attacks conditionally alter the model’s output only when certain input attributes are present, such as a specific speaker gender or spoken language. Inputs that do not meet the targeted condition are left unaffected, allowing the attack to precisely control when and how the model's behaviour is changed. This capability poses a serious risk, as it allows adversaries to manipulate or suppress responses based on sensitive user characteristics. Such a targeted adversarial attack is more difficult to detect and can exacerbate harmful biases, raising significant ethical and security concerns for the deployment of speech LLMs in real-world applications.

Through extensive experiments, we demonstrate that both general and selective attacks can be highly effective against speech LLMs. 
Our findings highlight the need for stronger defense strategies and more robust training approaches. 


\section{Previous Work}

\paragraph{Early Audio Attacks.}
Early research on adversarial attacks in spoken language processing (SLP) systems focused primarily on task-specific models, where the pipeline typically consisted of an ASR model followed by downstream components for tasks such as speech translation and intent classification~\citep{Tur2011SpokenLU, 1561278, ruan20-interspeech}. Initial work tended to focus on degrading ASR performance by maximizing word error rate (WER), using gradient-based perturbations to alter acoustic features and disrupt transcription accuracy~\citep{DBLP:journals/corr/abs-1711-03280, cisse2017houdini}.

Subsequent efforts shifted toward more realistic adversarial objectives by attempting \textit{targeted attacks}, where the adversary aims for a specific output transcription, such as injecting a malicious command like ``open evil.com” into the ASR output~\citep{yuan2018commandersong, DBLP:journals/corr/abs-1801-01944, DBLP:journals/corr/abs-1805-11852, qin2019imperceptible}. These attacks revealed the potential for manipulating ASR systems in ways that remained imperceptible to human listeners, using psychoacoustic models and frequency masking to hide perturbations~\citep{schönherr2018adversarial, Schnherr2018AdversarialAA}.
More recently, with the emergence of end-to-end speech language models, more real-world adversarial attack objectives have been explored, including for example safety jailbreaking~\citep{hughesattacking, roh2025multilingual, xiao2025tune, gupta2025bad}.

\paragraph{Universal Audio Attacks.}
Many of the above approaches require generating a separate adversarial perturbation for each input audio clip; but this is impractical in real-world, large-scale settings. To address this, researchers introduced universal adversarial perturbations (UAPs)—input-agnostic noise that could be applied broadly to many audio samples while achieving the adversary’s objective~\citep{DBLP:journals/corr/abs-1905-03828}. These universal perturbations typically required precise alignment with the original audio, limiting their usability in asynchronous or streaming environments. More recent work has proposed desynchronised UAPs, where the adversarial signal can be added at arbitrary points in the audio stream~\citep{10.1145-3372297.3423348} (e.g., pre-pending the attack signal to the audio sample, as is adopted in this work). These methods demonstrated a more practical route toward real-world attacks. These universal attacks have been applied to various older end-to-end architectures, including Listen-Attend-Spell (LAS), Connectionist Temporal Classification (CTC), and RNN-Transducer (RNN-T) models~\citep{lu2021exploring, raina20-interspeech}. The introduction of multi-task and instruction-following speech models, such as Whisper~\citep{radford2023robust}, gave the opportunity for novel adversarial objectives~\citep{raina2024muting, raina2024controlling}. 

Despite their improved practicality, existing universal attacks are typically unconstrained: the perturbation applies indiscriminately to any input. In contrast, this work is the first to propose \textit{selective universal attacks}, where the perturbation is only active for inputs that satisfy certain conditions, such as speaker gender/language. 
Beyond introducing selective attacks, we also analyse the threat posed by generalised adversarial attacks (e.g., muting and task control) on highly flexible speech-language systems that are receiving increasing attention.

\section{Adversarial Attack on SpeechLLM}


\subsection{Speech LLM}
\label{sec:speechllm}
A speech LLM is a system that integrates a pre-trained speech encoder with a generative language model, enabling it to perform a wide range of spoken language understanding and generation tasks. Given an input speech $\mathbf{x}=x_{1:N}$ consisting of $N$ frames, the speech encoder produces a sequence of hidden representations, $\mathbf{h}=\text{Enc}(\mathbf{x})$.
A transformation module (e.g., pooling layer~\cite{fathullah2024prompting}, Q-former~\cite{yu2024connecting}) further adapts the speech features to the LLM input space.
Let $\mathcal{P}_{\tt src}$ be the text prompt that instructs the model to perform a specific task, the speech LLM generates an output sequence $\hat{\bf y}$ with
\begin{equation}
{\hat{\bf y}} = \arg\max_{\bf y}{\tt P}({\bf y} \mid \text{Enc}(\mathbf{x}), \mathcal{P}_{\tt src})
\end{equation}

\subsection{Threat Model}

In this work, we consider a practical deployment scenario where the adversary lacks access to the internal architecture of the deployed speech language model and cannot directly alter the pre-defined task prompt, $\mathcal{P}_{\tt src}$. 
This reflects common real-world constraints, where users interact with models through restricted interfaces that expose only audio input and conceal underlying architecture or prompt details.
However, the adversary has access to the input audio and can therefore manipulate the speech signal at the acoustic level. Under these constraints, we employ a prepend-based \textit{universal} attack, where a short, fixed adversarial audio segment $\mathbf{a}$ is added to the beginning of any input utterance $\mathbf{x}$. This design eliminates the need for crafting signal-specific perturbations for different input audio, making it well-suited for real-time and low-latency applications.
To learn the adversarial segment, we assume white-box access to the speech LLM parameters that enables gradient-based optimisation. At inference time, the learned segment is expected to generalise to unseen speech inputs and remain effective across diverse settings. 

\subsection{General Attack}

The general attack learns an adversarial audio segment that, when prepended to any input audio, indiscriminately alters the model’s output regardless of the audio content or speaker identity.

\paragraph{Mute Outputs}

To perform a general muting attack, we aim to learn a short adversarial audio segment \( \mathbf{a} \) that, when prepended to any input speech \( \mathbf{x} \), causes the model to produce no output. The concatenated audio signal is denoted as \( \mathbf{a} \oplus \mathbf{x} \). The muting effect is achieved by encouraging the model to generate the special end-of-transcription token ($\mathtt{eot}$) as the first output, thereby terminating the output generation before any meaningful content is produced. The optimal adversarial segment \( \hat{\mathbf{a}} \) is learned via gradient-based optimisation on a small amount of speech data to reliably induce early termination across diverse inputs.
\begin{equation} \label{eqn:mute}
\hat{\mathbf{a}}=\arg\max_{\mathbf{a}}{\tt P}\left(y_1=\mathtt{eot} \mid \text{Enc}(\mathbf{a} \oplus \mathbf{x}), \mathcal{P}_{\tt src}\right)
\end{equation}
The obtained audio $\hat{\mathbf{a}}$ can also be interpreted as the acoustic realisation of the $\mathtt{eot}$ token~\cite{raina2024muting}. While $\mathcal{P}_{\tt src}$ is used during training to learn $\hat{\mathbf{a}}$, different prompts are provided during evaluation to assess the generalisation of the attack.


\paragraph{Task Control}
\label{sec:control}
In this setting, the goal is to manipulate the model’s behaviour so that it ignores the source prompt $\mathcal{P}_{\tt src}$ and instead performs a target task, $\mathcal{P}_{\tt tgt}$. We explore two training strategies with different learning targets: \texttt{Attack-ref}, which uses the reference transcriptions $\mathbf{y}_\text{tgt}$, and \texttt{Attack-hyp}, which relies on the model-generated hypotheses ${\hat{\mathbf{y}}}_\text{tgt}$ for the target task.

In \texttt{Attack-ref}, the audio attack is trained using, 
\begin{equation}
\hat{\mathbf{a}} = \arg\max_{\mathbf{a}} \, \tt{P} (\mathbf{y}_\text{tgt} \mid \text{Enc}(\mathbf{a} \oplus \mathbf{x}), \mathcal{P}_{\tt src})
\end{equation}
Alternatively in \texttt{Attack-hyp}, the hypothesis is obtained from
\begin{equation}
{\hat{\mathbf{y}}}_\text{tgt} = \arg\max_{\mathbf{y}} \, {\tt P}\left( \mathbf{y} \mid \text{Enc}(\mathbf{x}), \mathcal{P}_{\tt tgt} \right)
\end{equation}
where $\mathcal{P}_{\tt tgt}$ is a suitable prompt for the target task. This gives,
\begin{equation}
\hat{\mathbf{a}} = \arg\max_{\mathbf{a}} \, {{\tt P}} ({\hat{\mathbf{y}}}_\text{tgt} \mid \text{Enc}(\mathbf{a} \oplus \mathbf{x}), \mathcal{P}_{\tt src})
\end{equation}

\texttt{Attack-ref} uses manual reference outputs, allowing the attack to be trained with high-quality targets for effective control. However, collecting reference data may not always be feasible in practice. In contrast, \texttt{Attack-hyp} avoids the need for human annotations by using the model’s own predictions under the target prompt as pseudo-labels. 
Since the generated targets are inherently more aligned with the model's output style and distribution, this practice can make training easier.
In the experimental sections, we report and compare attack outcomes under both settings.

\subsection{Selective Attack}
\label{sec:select}
For a selective attack, the goal is to suppress the model’s output only when a particular attribute is present, while preserving the original outputs for all other inputs. We opt to demonstrate this attack on the muting adversarial objective (Equation \ref{eqn:mute}).

Let $f(\mathbf{x}) \in \{0, 1\}$ be an attribute function indicating whether the input audio $\mathbf{x}$ satisfies a specific condition (e.g., female speaker). The model's desired output ${\hat{\mathbf{y}}}_{\tt tgt}$ under a selective attack is:
\begin{equation}
\label{eq:select}
\small
\begin{aligned}
{\hat{\mathbf{y}}}_{\tt tgt} = 
&\begin{cases}
\mathtt{eot}, & \text{if } f(\mathbf{x}) = 1 \\
\arg\max_{\mathbf{y}} \, {\tt P} ( \mathbf{y} \mid \text{Enc}(\mathbf{x}), \mathcal{P}_{\tt tgt}), & \text{if } f(\mathbf{x}) = 0
\end{cases}
\end{aligned}
\end{equation}
Then the adversarial segment $\mathbf{a}$ is learned via
\begin{equation}
\hat{\mathbf{a}} = \arg\max_{\mathbf{a}} \, {\tt P} ({\hat{\mathbf{y}}}_{\tt tgt} \mid \text{Enc}(\mathbf{a} \oplus {\mathbf{x}}),\mathcal{P}_{\tt  src})
\end{equation}

In our setup, $\mathcal{P}_{\tt  tgt}$ is set to be the same as $\mathcal{P}_{\tt  src}$. This form of attack is more challenging than the general case because the adversarial segment must conditionally influence the model's output based on attributes of the input audio, without explicit access to those attributes at inference time. It needs to suppress outputs only when the target attribute is present, while preserving normal behaviour for all other inputs.

\subsection{Evaluation Metrics}

To evaluate the success of muting the model's outputs, we adopt two metrics following~\citet{raina2024muting}: the percentage of outputs that are empty or only consisting of blank tokens ($\varnothing$)~\footnote{Our calculation of $\varnothing$ differs slightly from that of~\citet{raina2024muting}, as we also consider blank sequences such as ``\textbackslash n\textbackslash n\textbackslash n'' or ``\quad\quad'' to be successful muted. Nonetheless, the resulting metrics are similar.}, and the average sequence length (\texttt{asl}) of the generated outputs. A higher $\varnothing$ value approaching 100\% and a \texttt{asl} close to 0 indicate a more successful attack in suppressing the model’s output.

Additionally, we use word error rate (WER) to evaluate the impact of the adversarial attack on the speech recognition performance. For the gender detection task, we report the classification accuracy. To assess whether the attack influences the language used in the model's output, we compute the average detected language probability $\tt P(lang)$ on the test set, using the open-sourced LangDetect~\cite{nakatani2010langdetect} toolkit.


\section{Experimental Setup}

\subsection{Models}

We study audio adversarial attacks on two speech LLMs: Qwen2-Audio \cite{chu2024qwen2} and Granite-Speech \cite{granite_speech_2025}.

\paragraph{Qwen2-Audio}~\cite{chu2024qwen2} integrates the Whisper large-v3 encoder~\cite{radford2023robust} with the Qwen-7B language model~\cite{bai2023qwen}, enabling the system to handle complex spoken language processing tasks. To bridge the modalities, the output of the Whisper encoder is downsampled via a pooling layer with a stride of two and mapped into the LLM's embedding space. The model undergoes a three-stage alignment process: multi-task pretraining, supervised fine-tuning for instruction-following, and direct preference optimisation to better align with human feedback. Qwen2-Audio achieves state-of-the-art performance across a range of audio processing tasks and has received significant attention \cite{wang2024audiobench, florea2025exploring, wu2025distinct}. Given its popularity and wide adoption, this paper investigates the potential risks and vulnerabilities associated with adversarial attacks on the model. In this paper, we experimented with the \texttt{Qwen2-Audio-7B-Instruct} model.

\paragraph{Granite-Speech} \citet{granite_speech_2025} extends the granite-3.2-8b-instruct model~\cite{granite2024granite} by incorporating a speech encoder composed of 10 Conformer blocks. The model follows the Q-former architecture~\cite{yu2024connecting} to transform the speech encoder outputs, where the outputs are downsampled by a factor of 5 using 3 trainable query vectors for every 15 speech embeddings. Granite-Speech is specifically designed for ASR on English and speech translation from English to other languages. It is trained on both publicly available datasets and synthetic data for the speech translation task. The \texttt{granite-speech-3.2-8b} model version is evaluated in this paper.

\subsection{Datasets}

We conduct experiments on two public speech datasets: LibriSpeech~\cite{panayotov2015librispeech} and FLEURS~\cite{conneau2023fleurs}. For setups on LibriSpeech, the dev\_other and test\_other sets are used in the attack learning and evaluation.
For FLEURS, we use ASR data from 5 languages: English (en), French (fr), German (de), Japanese (ja), and Chinese (zh). To assess the transferability of the learned attack, we further evaluate on three standard datasets: TED-LIUM3~\cite{hernandez2018ted}, MGB3 dev17b~\cite{bell2015mgb}, and the Artie bias corpus~\cite{meyer2020artie}, which we refer to as TED, MGB, and Artie.
Detailed information about all datasets is provided in Appendix~\ref{app:dataset}.

\subsection{Experimental Configuration}
During learning of the audio attack segment, the model weights of the speech LLMs are kept frozen, and only the learnable audio segment is updated via gradient descent. 
For the general attack and gender-based selective attack, we use the LibriSpeech dev\_other subset to learn the attack and evaluate on the test\_other subset, following the setup in \cite{raina2024muting}. For the language-based selective attack, we evaluate four configurations in which different language pairs are either retained or muted.
The models are trained on the combined FLEURS training set by merging the training data from both languages, with training targets defined according to Equation~\ref{eq:select}. Evaluation is conducted on the corresponding test set from two languages.
Unless otherwise specified, the attack segment length is set to 3.2 seconds, equivalent to 51,200 audio sample points. Hyperparameter settings used during training are given in Appendix \ref{sec:setup}. 

For decoding, we study both greedy search and beam search with a beam size of 5, with sampling disabled. Post-processing rules are applied to remove standard phrases generated by speech LLMs, such as \textit{``The original content of this audio is:''}. When computing WER, we use Whisper’s text normalisation scripts~\cite{radford2023robust} on both system outputs and ground-truth references.

\subsection{Prompts for Learning and Evaluation}

In our experiments, the system prompt is set to ``You are a helpful assistant.'' The text prompts used during training and evaluation are provided in Table~\ref{tab:prompt}, covering tasks such as ASR, speech translation, and gender detection. For brevity, we use abbreviated prompt names throughout the paper.

\begin{table}[!htbp]
    \centering
    \small
    \renewcommand\tabcolsep{2.2pt}
    \begin{tabularx}{\linewidth}{@{}l|X@{}}
    \toprule
         & Prompt \\
    \midrule
        $\mathcal{P}_\text{asr}$ & Transcribe the speech. \\
        \midrule
        $\mathcal{P}_\text{st-fr}$ & Translate the speech into French. \\
        $\mathcal{P}_\text{st-zh}$ & Translate the speech into Chinese. \\
    \midrule
        $\mathcal{P}_\text{gdr}$ & Detect the gender of the speaker. \\
        $\mathcal{P}_\text{gdr-fr}$ & Detect the gender of the speaker and reply in French. \\
        $\mathcal{P}_\text{gdr-zh}$ & Detect the gender of the speaker and reply in Chinese. \\
    \bottomrule
    \end{tabularx}
    \caption{List of text prompts and their abbreviations.}
    \label{tab:prompt}
\end{table}

\section{Results}
\subsection{General Attacks}

\subsubsection{Muting Outputs}


\begin{table}[!htbp]
    \centering
    \footnotesize
    \renewcommand\tabcolsep{4.5pt}
    \begin{tabular}{@{ }l|l|ccc|ccc}
    \toprule
    \multicolumn{2}{l|}{\multirow{2}*{Decode}} & \multicolumn{3}{c|}{Greedy Search} & \multicolumn{3}{c}{Beam Search} \\
    \multicolumn{2}{l|}{} & $\varnothing$ & \texttt{asl} & WER & $\varnothing$ & \texttt{asl} & WER \\
    \midrule
    \midrule
    \rowcolor{Gray}
    \multicolumn{8}{c}{Qwen2-Audio-7B} \\ 
    \multicolumn{2}{c|}{No Attack} & 0.0 & 17.6 & 6.6 & 0.0 & 23.8 & 5.2 \\
    \midrule
       \multirow{4}*{\rotatebox{90}{+Attack}} & \multirow{1}*{0.64s} & 0.0 & 16.3 & 22.1 & 0.0 & 17.2 & 14.1 \\
        & \multirow{1}*{1.6s} & 95.5 & 0.3 & 98.9 & 95.8 & 0.5 & 97.8 \\
        & \multirow{1}*{3.2s} & 100.0 & 0.0 & 100.0 & 99.9 & 0.0 & 100.0 \\
        & \multirow{1}*{6.4s} & 100.0 & 0.0 & 100.0 & 100.0 & 0.0 & 100.0 \\
     \midrule
     \midrule
     \rowcolor{Gray}
    \multicolumn{8}{c}{Granite-Speech-8B} \\ 
    \multicolumn{2}{c|}{No Attack} & 0.0 & 17.8 & 3.5 & 0.0 & 17.8 & 3.3 \\
    \midrule
       \multirow{4}*{\rotatebox{90}{+Attack}} & \multirow{1}*{0.64s} & 0.1 & 17.7 & 5.3 & 0.0 & 18.7 & 10.7 \\
        & \multirow{1}*{1.6s} & 74.0 & 5.9 & 68.7 & 17.3 & 17.5 & 69.4 \\
        & \multirow{1}*{3.2s} & 98.5 & 0.7 & 96.0 & 46.0 & 16.2 & 58.4 \\
        & \multirow{1}*{6.4s} & 100.0 & 0.0 & 99.9 & 81.1 & 7.7 & 74.6 \\
    \bottomrule
    \end{tabular}
    \caption{Percentage of muted samples ($\varnothing$), average sequence length (\texttt{asl}), and WER on LibriSpeech test\_other. Greedy and beam search results are shown for the no-attack baseline and universal adversarial attacks with prepended audio segments of varying lengths.}
    \label{tab:size}
\end{table}

Table \ref{tab:size} summarizes the results of the mute-all attack on Qwen2-Audio and Granite-Speech with different attack lengths. We aim to make the speech LLM generate nothing with the learned audio segment. During both training and evaluation, the text prompt $\mathcal{P}_\text{asr}$ is given to the model. We evaluate the attack performance using both greedy search and the more demanding setting of beam search using a beam width of 5. These results highlight the challenge of attacking a speech LLM. In prior work of attacking Whisper models \cite{raina2024muting}, the success rates dropped with larger model variants, and a 0.64-second adversarial audio segment was sufficient to mute the Whisper medium model (769M). In contrast, the models evaluated in our study are 9 times larger than Whisper, and thus require longer adversarial segments to achieve effective muting. As the results shown, a 0.64-second segment fails to mute the Qwen2-Audio model. However, extending the segment length to 3.2 seconds increases the mute success rate to near 100\% in both decoding settings. 


For the Granite-Speech model, using a 3.2-second adversarial segment, the mute-all attack achieves a 98.5\% success rate under greedy search. However, its effectiveness decreases when beam search is employed. With this setup, we adopt the default decoding parameters with the length penalty set to 1.0, which favours longer sequences. This tendency hinders the mute-all attack and contributes to the performance discrepancy. When the model is not fully muted, it occasionally hallucinates and repeats certain phrases during decoding.

In the following study, we will focus on attacking Qwen2-Audio, which receives more popularity and shows more vulnerability to the acoustic adversarial attacks. The attack segment length is set to 3.2 seconds, as this configuration showed strong performance in our evaluations. Results for beam search with a beam size of 5 are reported.



\begin{table}[!htbp]
    \centering
    \small
    \begin{tabular}{l|l|c|ccc}
    \toprule
     Eval & Metric & LBS & TED & MGB & Artie \\
    \midrule
     \multirow{2}*{$\mathcal{P}_\text{asr}$} & $\varnothing$ & 99.9 & 96.7 & 98.0 & 99.2 \\
     & \texttt{asl} & 0.0 & 0.1 & 0.1 & 0.1 \\
    \midrule
    \midrule
     \multirow{2}*{$\mathcal{P}_\text{st-fr}$} & $\varnothing$ & 99.7 & 97.0 & 98.0 & 99.2 \\
     & \texttt{asl} & 0.0 & 0.2 & 0.2 & 0.1 \\
    \midrule
     \multirow{2}*{$\mathcal{P}_\text{gdr}$} & $\varnothing$ & 97.0 & 85.2 & 97.3 & 97.0 \\
     & \texttt{asl} & 0.0 & 0.6 & 0.1 & 0.1 \\
    \bottomrule
    \end{tabular}
    \caption{Attack transferability across datasets and prompts. The 3.2s adversarial segment trained on LibriSpeech is evaluated on other ASR test sets. We report the percentage of muted samples ($\varnothing$) and average sequence length (\texttt{asl}). While the attack is learned with the prompt $\mathcal{P}_\text{asr}$, two additional prompts, $\mathcal{P}_\text{st-fr}$ and $\mathcal{P}_\text{gdr}$, are evaluated to assess its generalisation ability.}
    \label{tab:trans}
\end{table}


Transferability is a key aspect of adversarial attacks, particularly in the context of speech LLMs that are designed to operate across diverse inputs and tasks. Tables \ref{tab:trans} and \ref{tab:trans_fleurs} present an evaluation of the transferability of the mute-all audio attack learned on LibriSpeech across three other English ASR datasets, as well as FLEURS test sets spanning four additional languages. Firstly, we evaluate the attack using the same $\mathcal{P}_\text{asr}$ prompt that was used during training. This results in over 96\% success rates on all English test sets and achieves consistently strong performance across most other languages, except for German (74.0\%). These results suggest that the learned adversarial audio generalises well across datasets and languages.


\begin{table}[!htbp]
    \centering
    \renewcommand\tabcolsep{5pt}
    \small
    \begin{tabular}{l|l|ccccc}
    \toprule
     Eval & Metric & en & fr & de & ja & zh \\
    \midrule
     \multirow{2}*{$\mathcal{P}_\text{asr}$} & $\varnothing$ & 100.0 & 92.0 & 74.0 & 93.7 & 98.4 \\
     & \texttt{asl} & 0.0 & 1.1 & 4.3 & 3.5 & 1.1 \\
    \midrule
    \midrule
     \multirow{2}*{$\mathcal{P}_\text{st-fr}$} & $\varnothing$ & 100.0 & 92.0 & 58.7 & 88.9 & 99.8 \\
     & \texttt{asl} & 0.0 & 1.1 & 6.0 & 6.5 & 1.9 \\
    \midrule
     \multirow{2}*{$\mathcal{P}_\text{gdr}$} & $\varnothing$ & 100.0 & 94.8 & 83.3 & 92.6 & 77.9 \\
     & \texttt{asl} & 0.0 & 0.2 & 0.9 & 3.3 & 0.5 \\
    \bottomrule
    \end{tabular}
    \caption{Attack transferability across datasets and prompts. The 3.2s adversarial segment trained on English data from LibriSpeech is evaluated on five ASR test sets from FLEURS, covering different languages.}
    \label{tab:trans_fleurs}
\end{table}

Given that speech LLMs can adapt to a wide range of tasks via prompt modification, we further assess whether the adversarial attack remains effective under different task prompts. Specifically, we evaluate with the prompt $\mathcal{P}_\text{st-fr}$ and $\mathcal{P}_\text{gdr}$ that instruct the model for speech translation into French and gender detection tasks. Although the given text prompt differs from the one used during training, the mute-all attack still achieves a success rate of over 97\% on the English test sets. Moreover, the attack remains effective on speech inputs in other languages on FLEURS.
This highlights the robust transferability of the adversarial audio across different task settings, reinforcing the generalisability and threat potential of such universal attacks.

\subsubsection{Task Control}

\begin{table}[!htbp]
    \centering
    \small
    \begin{tabular}{l|l|c|ccc}
    \toprule
       & Eval & WER & $\tt P(en)$ & $\tt P(fr)$ & $\tt P(zh)$ \\
    \midrule
       \multirow{3}*{No Attack} & $\mathcal{P}_\text{asr}$ & 5.2 & 96.8 & 0.4 & 0.1 \\
       & $\mathcal{P}_\text{st-fr}$ & 89.2 & 17.5 & 80.8 & 0.2 \\
       & $\mathcal{P}_\text{st-zh}$ & 66.4 & 32.0 & 0.1 & 63.0 \\
    \midrule
    \midrule
        \multirow{2}*{Attack-hyp} & $\mathcal{P}_\text{st-fr}$ & 6.3 & 96.0 & 0.2 & 0.0 \\
        & $\mathcal{P}_\text{st-zh}$ & 6.4 & 96.0 & 0.0 & 0.0 \\
        \midrule
        \multirow{2}*{Attack-ref} & $\mathcal{P}_\text{st-fr}$ & 7.2 & 96.4 & 0.8 & 0.0 \\
        & $\mathcal{P}_\text{st-zh}$ & 7.1 & 96.2 & 0.3 & 0.6 \\
    
    \bottomrule
    \end{tabular}
    \caption{Task control attack results on LibriSpeech test\_other. In the training, text prompt for French translation ($\mathcal{P}_\text{st-fr}$) is given, while the adversarial segment redirects it to perform English ASR. WER and detected language probabilities are reported. \texttt{Attack-hyp} uses the model’s own ASR predictions as learning targets, while \texttt{Attack-ref} is learned with manual references.}
    \label{tab:task}
\end{table}

Table \ref{tab:task} presents the results of a different class of general adversarial attacks: task control. We provide the LLM with the $\mathcal{P}_\text{st-fr}$ prompt, which is intended for speech translation into French, while attacking the model into producing English ASR outputs. This creates a deliberate mismatch between the prompt and the expected task, allowing us to assess the model’s vulnerability to task reprogramming. Here, we explore two experiment settings with different training targets: \texttt{Attack-hyp} and \texttt{Attack-ref} as discussed in Section~\ref{sec:control}. 


Given the $\mathcal{P}_\text{st-fr}$ prompt, for the baseline setup without adversarial attacks, the model follows the instruction and generates 80.8\% responses in French. Nevertheless, the audio attack redirects the model to output more than 96.0\% contents in English under both training setups. When comparing the generated outputs to the ASR references, similar WER results (7.2 and 6.3) to the baseline of prompting with $\mathcal{P}_\text{asr}$ (5.2) can be observed, indicating the high quality of the generated transcriptions. Additionally, we test the attack under a new prompt, $\mathcal{P}_\text{st-zh}$ that is not present in the training, and observe comparable performance to the $\mathcal{P}_\text{st-fr}$ setup, indicating great transferability of the attack.

\begin{table}[!htbp]
    \centering
    \small
    \begin{tabular}{l|l|c|ccc}
    \toprule
        & Eval & Acc & $\tt P(en)$ & $\tt P(fr)$ & $\tt P(zh)$ \\
    \midrule
       \multirow{3}*{No Attack} & $\mathcal{P}_\text{gdr}$ & 92.7 & 100.0 & 0.0 & 0.0 \\
       & $\mathcal{P}_\text{gdr-fr}$ & 91.7 & 17.8 & 82.2 & 0.0 \\
       & $\mathcal{P}_\text{gdr-zh}$ & 92.7 & 12.9 & 0.0 & 87.0 \\
    \midrule
    \midrule
    \multirow{3}*{Attack-hyp} & $\mathcal{P}_\text{gdr}$ & 90.7 & 100.0 & 0.0 & 0.0 \\
       & $\mathcal{P}_\text{gdr-fr}$ & 91.5 & 95.3 & 4.7 & 0.0 \\
       & $\mathcal{P}_\text{gdr-zh}$ & 89.7 & 57.6 & 0.0 & 42.4 \\
    \midrule
    \multirow{3}*{Attack-ref} & $\mathcal{P}_\text{gdr}$ & 60.8 & 99.4 & 0.0 & 0.0 \\
       & $\mathcal{P}_\text{gdr-fr}$ & 55.6 & 95.9 & 1.3 & 0.0 \\
       & $\mathcal{P}_\text{gdr-zh}$ & 66.6 & 62.0 & 0.1 & 34.9 \\
    \bottomrule
    \end{tabular}
    \caption{Transferability of the task control attack for gender detection on LibriSpeech test\_other. The attack is trained to induce English ASR outputs and evaluated using prompts for gender classification with multilingual response generation. Gender classification accuracy and detected language probabilities are reported.}
    \label{tab:transfer}
\end{table}

We further evaluate the model’s capability to perform other tasks under attack, as shown in Table~\ref{tab:transfer}. Specifically, we prompt the model to identify the speaker’s gender and respond in different languages. We report both the accuracy of the gender classification task and the language probability of the generated responses. 
Under the \texttt{Attack-hyp} setup where the model is trained using its own ASR decoding outputs, the gender detection performance remains comparable to the baseline (non-attacked) setting, indicating that the adversarial attack does not impair this aspect of the model’s reasoning. Nonetheless, the attack segment affects the language generation behaviour: despite being prompted with $\mathcal{P}_\text{gdr-fr}$ or $\mathcal{P}_\text{gdr-zh}$, the model predominantly replies in English.
In contrast, different behaviours are observed under the \texttt{Attack-ref} setup, where training is based on manual references. In many cases, the model outputs English ASR transcriptions instead of performing gender detection, leading to a decrease in accuracy. These findings suggest that attacks trained on self-generated hypotheses are less disruptive to non-ASR tasks, whereas using cleaner reference targets may cause the model to overfit to the training task and rewrite its ability to perform other tasks.


\subsection{Selective Attacks}

In this section we consider \textit{selective universal attacks}, as introduced in Section \ref{sec:select}. As an initial demonstration, the experiments here focus on selecting based on two specific attributes of speech: speaker gender and spoken language. The models are prompted with $\mathcal{P}_\text{asr}$ in the learning and evaluation to perform speech recognition. 

\subsubsection{Gender-based Attack}

As shown in Table~\ref{tab:transfer}, Qwen2-Audio achieves zero-shot gender detection accuracy of 92.7\% on test\_other, indicating that the model is capable of capturing gender-related cues from speech. Leveraging this ability, we evaluate the selective attack conditioned on speaker gender. In the \texttt{Mute-female} setting, only speech samples from female speakers are muted, while the model is trained to transcribe all male speech. Conversely, in the \texttt{Mute-male} setting, the model is trained to output transcriptions normally for female speakers and outputs nothing for male speakers. 
For both the baseline and attack settings, we compute the WER separately for male and female speakers, along with metrics relevant to the muting attack. 

\begin{table}[!htbp]
    \centering
    \small
    \renewcommand\tabcolsep{3.8pt}
    \begin{tabular}{l|c|c|c|c|ccc}
    \toprule
      & Set & $\varnothing$ & \texttt{asl} & WER & ins & del & sub \\
    \midrule
      \multirow{2}*{No Attack} 
      & M & 0.0 & 16.5 & 5.9 & 0.8 & 1.0 & 4.1 \\
      & F & 0.0 & 19.1 & 4.5 & 0.4 & 0.9 & 3.2 \\
    \midrule
    \midrule
    \multirow{2}*{Mute-female} & M & 14.7 & 14.4 & 19.4 & 1.1 & 14.0 & 4.3  \\
    & F & 92.2 & 2.2 & 89.2 & 0.1 & 88.8 & 0.3  \\
    \midrule
    \multirow{2}*{Mute-male} & M & 71.2 & 4.8 & 73.2 & 0.4 & 71.5 & 1.3 \\
    & F & 0.3 & 18.9 & 6.6 & 0.6 & 2.1 & 4.0 \\
    \bottomrule
    \end{tabular}
    \caption{Gender-based selective attack results on LibriSpeech test\_other. ``M'' and ``F'' denote male and female speech samples. For each group, we report WER results and its composition, as well as the percentage of muted samples ($\varnothing$) and average sequence length (\texttt{asl}).}
    \label{tab:gender}
\end{table}

Table~\ref{tab:gender} presents the selective attack results under two settings. In the \texttt{Mute-female} setup, 92.2\% of female speeches are successfully suppressed, while some male speech samples (14.7\%) are incorrectly muted. This results in increases in the deletion error, particularly for the female group. For \texttt{Mute-male}, 71.2\% of male recordings are successfully muted with our proposed audio attack, while most female recordings are transcribed without being affected. These results indicate that the selective muting attack is effective at controlling the model's behaviours based on the gender attribute.

\begin{figure}[!htbp]
    \centering
    \includegraphics[width=0.88\linewidth]{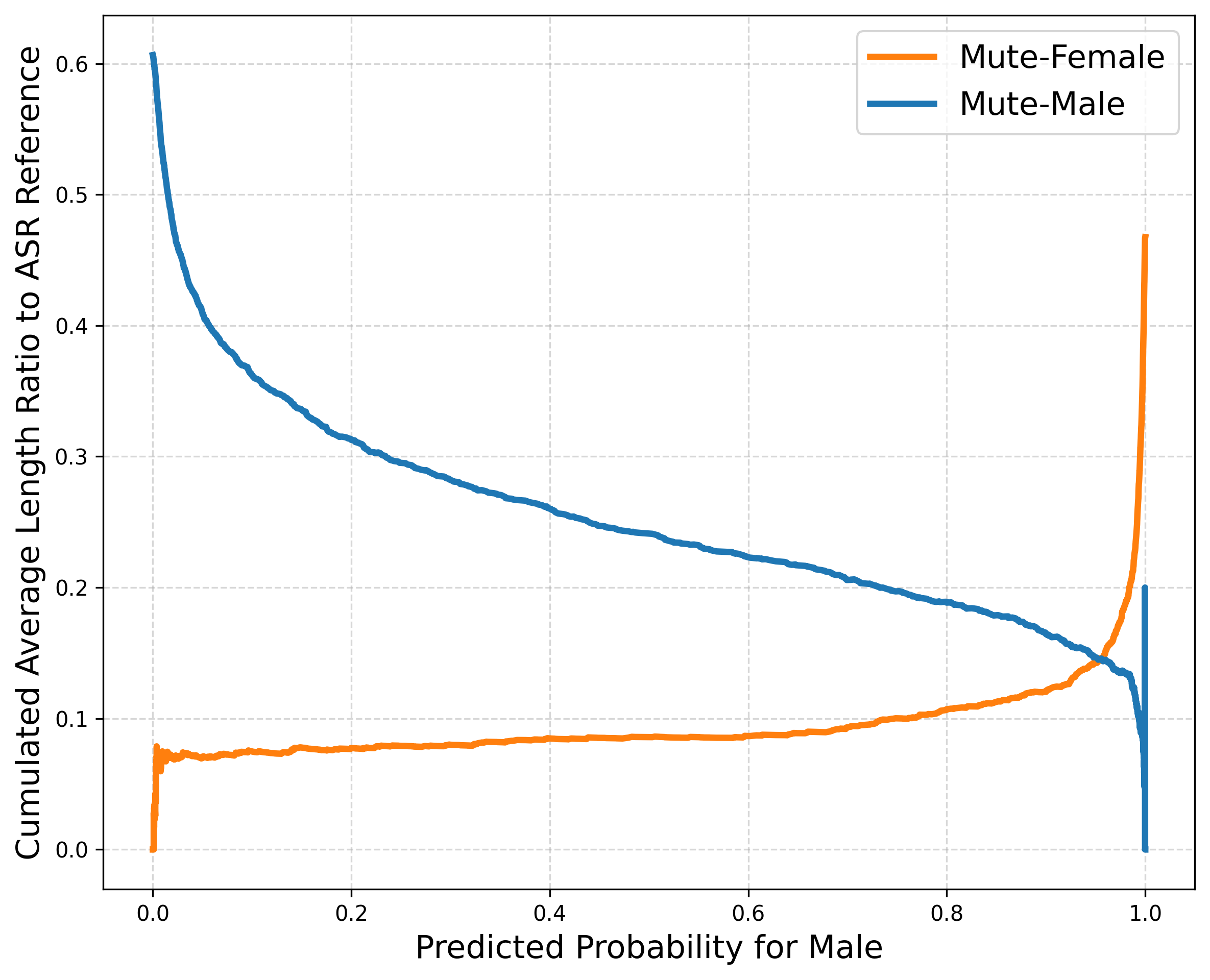}
    \caption{Cumulative average output length ratio (relative to ASR references) for \texttt{Mute-female} and \texttt{Mute-male} models, plotted against Qwen-Audio's zero-shot gender classification probabilities using $\mathcal{P}_{\text{gdr}}$.}
    \label{fig:cumm}
\end{figure}


In Figure~\ref{fig:cumm}, we further analyse how the model’s perception of gender influences its predictions, providing insight into the relationship between internal attribute representation and selective muting outcomes. 
We observe that the model's predicted probability for the male class is positively correlated with the length of the outputs in the \texttt{Mute-female} setup, and a similar trend can be observed in the \texttt{Mute-male} setup. These findings suggest that the model’s internal perception of gender is crucial for the effectiveness of the selective attack. 



\subsubsection{Language-based Attack}

\begin{table}[!htbp]
    \centering
    \small
    \renewcommand\tabcolsep{3.9pt}
    \begin{tabular}{l|c|c|c|c|ccc}
    \toprule
     Attack & Set & $\varnothing$ & \texttt{asl} & WER & ins & del & sub \\
    \midrule
      \multirow{3}*{No Attack} & en & 0.0 & 22.3 & 5.6 & 0.8 & 1.4 & 3.4 \\
      & fr & 0.0 & 25.1 & 10.3 & 1.7 & 1.8 & 6.8 \\
      & zh & 0.0 & 39.1 & 9.9 & 1.4 & 4.5 & 4.0 \\
    \midrule
    \midrule
    \multirow{2}*{Mute-en} & en & 85.4 & 3.4 & 90.6 & 4.4 & 84.3 & 1.8 \\
    & fr & 0.0 & 25.1 & 12.6 & 2.2 & 2.1 & 8.4 \\
    \midrule
    \multirow{2}*{Mute-fr} & en & 2.6 & 21.2 & 19.0 & 2.6 & 8.2 & 8.2 \\
    & fr & 62.1 & 7.0 & 89.7 & 1.0 & 73.9 & 14.8 \\
    \midrule
    \midrule
    \multirow{2}*{Mute-en} & en & 85.4 & 2.6 & 92.6 & 1.1 & 89.4 & 2.1 \\
    & zh & 0.0 & 38.7 & 10.3 & 1.2 & 4.6 & 4.5 \\
    \midrule
    \multirow{2}*{Mute-zh} & en & 0.0 & 22.3 & 14.9 & 2.2 & 3.1 & 9.6 \\
    & zh & 92.6 & 3.3 & 94.0 & 0.5 & 92.8 & 0.7 \\
    \bottomrule
    \end{tabular}
    \caption{Language-based selective attack results on FLEURS test sets for English (en), French (fr), and Chinese (zh). For each group, we report WER or CER and its composition, as well as the percentage of muted samples ($\varnothing$) and average sequence length (\texttt{asl}).}
    \label{tab:lang}
\end{table}

We further extend the selective attack framework to the language attribute, aiming to mute the model’s outputs for speech in a specific language while ensuring the model continues to generate speech transcriptions for others. 
We tested four setups, each involving a language pair: (1) keep French, mute English; (2) keep English, mute French; (3) keep Chinese, mute English; and (4) keep English, mute Chinese. 
As shown in Table~\ref{tab:lang}, the attack achieves high success rates in suppressing the targeted language without significantly affecting the model’s performance on the non-targeted language. Since English and Mandarin are more linguistically distinct than English and French, the selective attack achieves a higher success rate in this case, as it is easier for the model to differentiate between the two languages. Results demonstrate that the model encodes language information in a way that can be exploited by the universal audio prompt to conditionally disrupt outputs. 

\section{Conclusions}

In this paper, we explore the use of universal acoustic adversarial attacks against speech LLMs. For general attacks, it is demonstrated that universal audio segments can be designed to mute the model entirely or explicitly designed to override the textual prompt instruction, causing the model to perform a different (unintended) task. A success rate of over 95\% is observed in these attack scenarios, highlighting the effectiveness of the proposed methods. 
Beyond these general attacks, we also proposed a novel class of \textit{selective} universal acoustic attacks, in which the adversarial input is conditionally activated based on specific attributes of the input speech signal, such as gender or language. Our results show that these attacks can be carefully crafted to selectively suppress outputs for targeted groups, while leaving others unaffected. 
These findings raise important concerns about the safety, robustness, and fairness of speech LLMs, and underscore the need for developing more robust architectures and defense mechanisms in future work.

\section*{Acknowledgments}

This paper reports on research supported by Cambridge University Press \& Assessment, a department of The Chancellor, Masters, and Scholars of the University of Cambridge.

\section{Limitations}

While our study demonstrates the vulnerabilities in speech LLMs to universal and selective adversarial attacks, several limitations remain. First, our attacks are evaluated primarily on a popular speech LLM, Qwen2-Audio. It adopts a widely used architecture to combine a speech encoder with a general-purpose language model, reflecting the design of a prevalent class of speech LLMs. While we expect the findings to generalise to other models with similar architectures, further experiments across diverse models is needed to fully assess the effectiveness and limitations of these attacks. Second, the attack relies on access to the model’s weights during training (i.e., white-box settings). The performance of similar attacks in black-box scenarios remains to be explored and may differ significantly.

\section{Risks and Ethics}
The findings presented in this work expose vulnerabilities in speech LLMs, highlighting the potential for adversarial audio attacks to exploit these systems in harmful ways. The transferability and universality of the attacks suggest that a single adversarial audio input could affect diverse tasks and users across applications.
Selective attacks, in particular, pose a risk of reinforcing or introducing discriminatory behaviours in model outputs. For example, an adversary could intentionally mute or manipulate outputs only for specific demographic groups, leading to biased or unequal access to information and services. Such attacks may be difficult to detect, especially when applied in subtle or targeted ways. By highlighting these issues, we aim to encourage future research to focus on robust training methods and adversarial resilience, to enable the equitable and trustworthy deployment of speech-language systems.

\bibliography{custom}

\appendix

\section{Experimental Details}

\subsection{Dataset}
\label{app:dataset}

LibriSpeech~\cite{panayotov2015librispeech} is a corpus of read-aloud audiobook speech featuring a diverse range of speakers. In this work, we use the dev\_other and test\_other subsets that contain more challenging and acoustically diverse speech samples. For the gender-based selective attack, we leverage the gender annotations provided in the LibriSpeech metadata, which offers a relatively balanced distribution between male and female speakers.
FLEURS~\cite{conneau2023fleurs} is a multilingual dataset containing n-way parallel speech data in 102 languages, commonly used for ASR and speech translation tasks. We use the training and evaluation data from five selected languages in our experiments.
TED-LIUM3~\cite{hernandez2018ted} is a large-scale English speech corpus derived from TED talks. The MGB-3 dataset~\cite{bell2015mgb} consists of British television broadcast recordings spanning diverse genres, including news, drama, comedy, and documentaries. Lastly, the Artie bias dataset~\cite{meyer2020artie}, is sourced from the Common Voice project~\cite{ardila2020common} and comprises speech recordings collected from speakers with a wide range of demographic backgrounds. Detailed statistics are listed in Table \ref{tab:datas}.

\begin{table}[!htbp]
    \centering
    \small
    \begin{tabular}{l|c|ccc}
    \toprule
        Data & Split & \#Utts & \#Spkers & Hours \\
    \midrule
        \multirow{2}*{LibriSpeech} & dev\_other & 2864 & 91 & 5.1 \\
        & test\_other & 2939 & 90 & 5.3 \\
    \midrule
        \multirow{9}*{FLEURS} & train\_en & 2516 & 1464 & 7.3 \\
        & train\_fr & 3196 & 1503 & 10.3 \\
        & train\_zh & 3246 & 1500 & 9.7 \\
    \cmidrule{2-5}
        & test\_en & 645 & 349 & 1.8 \\
        & test\_fr & 676 & 332 & 1.9 \\
        & test\_de & 858 & 345 & 3.1 \\
        & test\_ja & 650 & 321 & 2.4 \\
        & test\_zh & 945 & 349 & 3.1 \\
    \midrule
        TED-LIUM3 & test & 1155 & 15 & 2.6 \\
        MGB3 & dev17b & 5856 & 173 & 4.6 \\
        Artie & test & 1712 & 969 & 2.4 \\
    \bottomrule
    \end{tabular}
    \caption{Statistics of the datasets used in this paper.}
    \label{tab:datas}
\end{table}

\subsection{Licensing}

This work is conducted on datasets that are either publicly available or authorised for research use, such as MGB-3. Our implementation is based on PyTorch 2.3.1, an open-source machine learning framework. The code and model weights for Qwen2-Audio and Granite-Speech are publicly released. We adhere to their terms of the Apache-2.0 license in all aspects of our usage. We ensured that the use of all existing datasets was consistent with their originally intended use as specified by their creators. 


\subsection{Training Setup}
\label{sec:setup}
All experiments are conducted on a single NVIDIA A100 GPU, using a batch size of 8. For the mute outputs attack and the task control attack, the acoustc attack segment is trained for 80 epochs, resulting in around 23 and 31 GPU hours, respectively. For the gender-based and the language-based selective attack, the acoustc attack segment is trained for 60 and 40 epochs, corresponding to 31 and 23 GPU hours, respectively. When learning the audio attack without an amplitude constraint, we set the learning rate to 1e-2; for experiments with the amplitude constraint, the learning rate is reduced to 1e-4 to stabilise the optimisation process. A cosine learning rate schedule with a weight decay of 0.01 is used throughout training.

\section{Results with Constrained Amplitude}

Imperceptibility is a critical property of adversarial attacks, particularly in the speech domain, where input signals are not only processed by models but also perceived by human listeners. For an attack to be practical and stealthy in real-world scenarios, it should sound natural and avoid arousing suspicion, making it more difficult to detect or filter. The adversarial audio segments produced by the proposed method in this paper resemble random noise, making them inherently difficult to detect. In this section, we further apply a stealth constraint to ensure that the adversarial segment remains imperceptible to human listeners. Specifically, we enforce an $\ell_\infty$ norm constraint on the adversarial audio segment, restricting the maximum absolute value of any waveform sample,
\begin{equation}
    \left\| \hat{\mathbf{a}} \right\|_\infty \leq \epsilon
\end{equation}
where $\hat{\mathbf{a}}$ denotes the learned adversarial audio segment. This constraint keeps the adversarial audio effectively undetectable to human listeners, while still successfully influencing the model's behaviour.

\begin{table}[!htbp]
    \centering
    \small
    \begin{tabular}{@{ }l|l|ccc}
    \toprule
    \multicolumn{2}{l|}{\multirow{2}*{Models}} & \multicolumn{3}{c}{Qwen2-Audio-7B} \\
    \multicolumn{2}{l|}{} & $\varnothing$ & \texttt{asl} & WER \\
    \midrule
    \midrule
        \multicolumn{2}{c|}{No Attack} & 0.0 & 23.8 & 5.2 \\
    \midrule
    \midrule
       \multirow{3}*{\rotatebox{90}{+Attack}} & \multirow{1}*{$\epsilon=0.005$} & 97.2 & 0.6 & 96.8 \\
        & \multirow{1}*{$\epsilon=0.02$} & 98.3 & 0.4 & 98.0 \\
        & \multirow{1}*{$\epsilon=\inf$} & 99.9 & 0.0 & 100.0 \\
    \bottomrule
    \end{tabular}
    \caption{Results of mute output attack with varying amplitude constraints ($\epsilon$) under the $\mathcal{P_\text{asr}}$ prompt. Percentage of muted samples ($\varnothing$), average sequence length (\texttt{asl}), and WER on LibriSpeech test\_other are reported.}
    \label{tab:size_eplison}
\end{table}

Table~\ref{tab:size_eplison} presents results for both the no-attack baseline and universal adversarial attacks using prepended audio segments under varying amplitude constraints ($\epsilon$) for the mute output attack. As the results indicate, limiting the imperceptibility of the adversarial audio does not significantly degrade the attack performance. Based on these findings, we set $\epsilon=0.02$ in the following experiments to investigate its impact on other attack scenarios.

\begin{figure}[!htbp]
    \centering
    \vspace{0.5em}
    \begin{subfigure}[b]{1.0\linewidth}
        \centering
        \includegraphics[width=\linewidth,trim={2.2cm 0.6cm 4cm 1.73cm},clip]{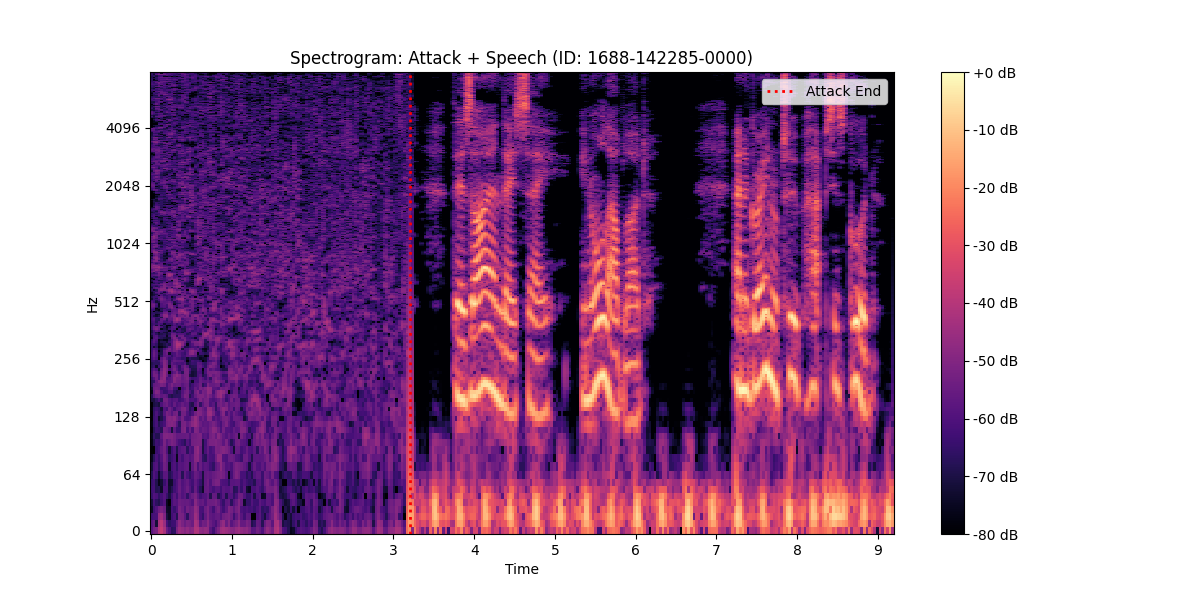}
        \caption{(a) $\epsilon=0.005$}  
    \end{subfigure}
    
    \vspace{1em}

    \begin{subfigure}[b]{1.0\linewidth}
        \centering
        \includegraphics[width=\linewidth,trim={2.2cm 0.6cm 4cm 1.73cm},clip]{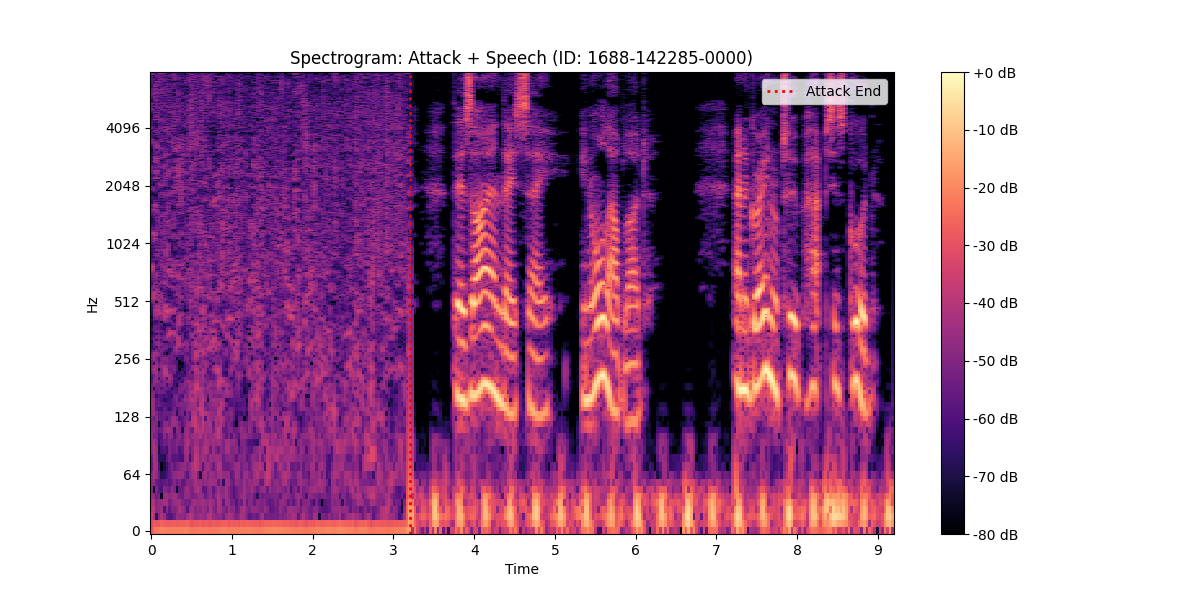}
        \caption{(a) $\epsilon=0.02$}  
    \end{subfigure}
    
    \vspace{1em}

    \begin{subfigure}[b]{1.0\linewidth}
        \centering
        \includegraphics[width=\linewidth,trim={2.2cm 0.6cm 4cm 1.73cm},clip]{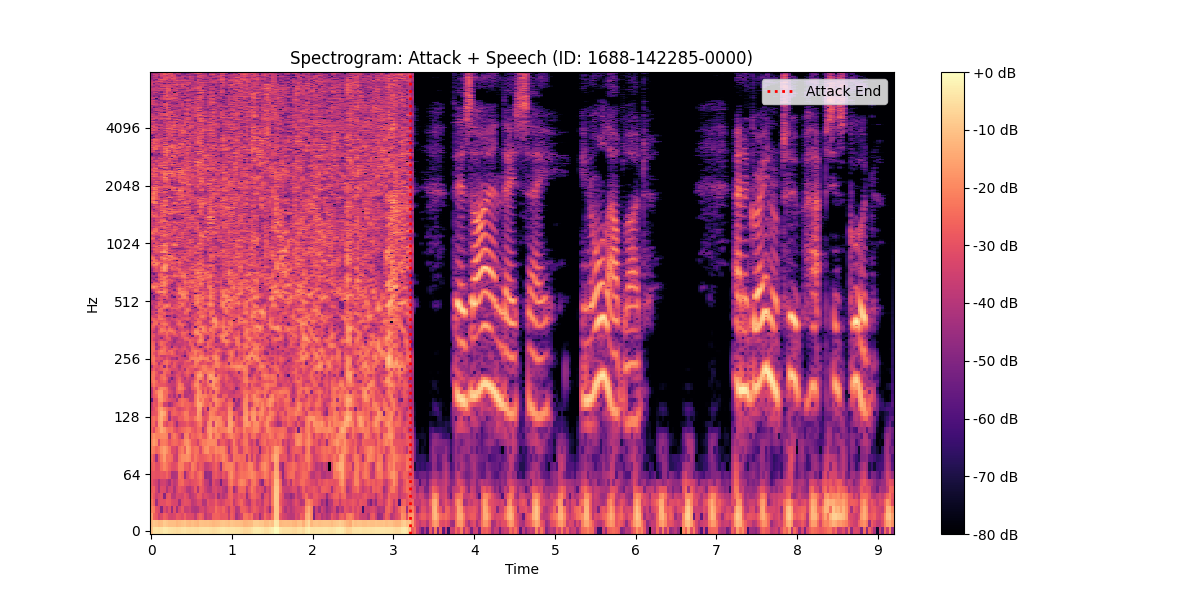}
        \caption{(b) $\epsilon=\inf$}
    \end{subfigure}

    \caption{Mel spectrograms of universal adversarial segments prepended to a sample from the test\_other set, shown for attacks with different amplitude constraints.}
    \label{fig:map}
\end{figure}

In Figure~\ref{fig:map}, we present Mel spectrogram plots of adversarial segments under different amplitude constraints. The spectrograms reveal clear structural differences in the learned audio signals. The segment with $\epsilon=\inf$ displays stronger and more concentrated energy patterns, indicating more aggressive perturbations that may be perceptible to human listeners. In contrast, at $\epsilon=0.02$, the adversarial signal appears visually sparse with low energy distributed across time and frequency. This is consistent with the objective of imperceptibility, as the perturbation more effectively adheres to the stealth constraint.


\begin{table}[!htbp]
    \centering
    \small
    \begin{tabular}{l|l|c|ccc}
    \toprule
        & Prompt & WER & $\tt P(en)$ & $\tt P(fr)$ & $\tt P(zh)$ \\
    \midrule
       \multirow{3}*{No Attack} & $\mathcal{P}_\text{gdr}$ & 92.7 & 100.0 & 0.0 & 0.0 \\
       & $\mathcal{P}_\text{gdr-fr}$ & 91.7 & 17.8 & 82.2 & 0.0 \\
       & $\mathcal{P}_\text{gdr-zh}$ & 92.7 & 12.9 & 0.0 & 87.0 \\
    \midrule
    \midrule
    \multirow{1}*{Attack-hyp} & $\mathcal{P}_\text{st-fr}$ & 7.2 & 96.4 & 0.8 & 0.0 \\
    ($\epsilon=\inf$)    & $\mathcal{P}_\text{st-zh}$ & 7.1 & 96.2 & 0.3 & 0.6 \\
    \midrule
    \multirow{1}*{Attack-hyp} & $\mathcal{P}_\text{st-fr}$ & 7.9 & 95.7 & 1.5 & 0.0 \\
    ($\epsilon=0.02$) & $\mathcal{P}_\text{st-zh}$ & 7.6 & 95.5 & 0.4 & 0.1 \\
    \bottomrule
    \end{tabular}
    \caption{Results of task control attack on LibriSpeech test\_other under the $\epsilon=\inf$ and $\epsilon=0.02$ constraints. Qwen2-Audio is prompted for French translation ($\mathcal{P}_\text{st-fr}$), while the adversarial segment redirects it to perform English ASR. WER and detected language probabilities are reported for the no-attack baseline and the \texttt{Attack-hyp} setting.}
    \label{tab:task_eplison}
\end{table}

As shown in Table~\ref{tab:task_eplison}, the task control attack achieves comparable results with $\epsilon=0.02$ to that with $\epsilon=\inf$. Under the $\mathcal{P}_\text{st-fr}$ setting, the WER increases slightly from 7.2 to 7.9, and the detected language probability for English, P(en), decreases from 96.4 to 95.7. 
These small differences suggest that even a small perturbation can be highly effective for the task control attack.


\begin{table}[!htbp]
    \centering
    \small
    \renewcommand\tabcolsep{3.8pt}
    \begin{tabular}{l|c|c|c|c|ccc}
    \toprule
     Attack & Set & $\varnothing$ & \texttt{asl} & WER & ins & del & sub \\
    \midrule
      \multirow{2}*{No Attack} 
      & M & 0.0 & 16.5 & 5.9 & 0.8 & 1.0 & 4.1 \\
      & F & 0.0 & 19.1 & 4.5 & 0.4 & 0.9 & 3.2 \\
    \midrule
    \midrule
    \multirow{2}*{Mute-female} & M & 13.3 & 14.5 & 20.1 & 1.7 & 13.8 & 4.6 \\
    & F & 85.4 & 3.7 & 83.8 & 1.0 & 81.7 & 1.2 \\
    \midrule
    \multirow{2}*{Mute-male} & M & 68.4 & 5.2 & 71.2 & 0.5 & 69.0 & 1.7 \\
    & F & 0.0 & 19.1 & 6.0 & 0.7 & 1.4 & 3.9 \\
    \bottomrule
    \end{tabular}
    \caption{Gender-based selective attack results on LibriSpeech test\_other under the $\epsilon=0.02$ constraint. ``M'' and ``F'' denote male and female speech samples. For each group, we report WER results and its composition, as well as the percentage of muted samples ($\varnothing$) and average sequence length (\texttt{asl}).}
    \label{tab:gender_eplison}
\end{table}

Table~\ref{tab:gender_eplison} shows the impact of the imperceptibility constraint on the gender-based selective attack. Compared to results in Table~\ref{tab:gender}, the percentage of muted samples for the targeted gender decreases (from 92.2\% to 85.4\% in the \texttt{Mute-female} setting and from 71.2\% to 68.4\% in the \texttt{Mute-male} setting). The results imply a trade-off between imperceptibility and attack effectiveness in the selective attack scenario.

\begin{table}[!htbp]
    \centering
    \small
    \renewcommand\tabcolsep{3.9pt}
    \begin{tabular}{l|c|c|c|c|ccc}
    \toprule
     Attack & Set & $\varnothing$ & \texttt{asl} & WER & ins & del & sub \\
    \midrule
      \multirow{3}*{No Attack} & en & 0.0 & 22.3 & 5.6 & 0.8 & 1.4 & 3.4 \\
      & fr & 0.0 & 25.1 & 10.3 & 1.7 & 1.8 & 6.8 \\
      & zh & 0.0 & 39.1 & 9.9 & 1.4 & 4.5 & 4.0  \\
    \midrule
    \midrule
    \multirow{2}*{Mute en} & en & 21.6 & 17.6 & 27.3 & 0.9 & 23.0 & 3.4 \\
    & fr & 0.0 & 25.1 & 14.0 & 2.1 & 2.0 & 9.8 \\
    \midrule
    \multirow{2}*{Mute fr} & en & 0.0 & 22.2 & 7.7 & 1.2 & 2.3 & 4.2 \\
    & fr & 32.5 & 16.8 & 58.4 & 3.8 & 36.3 & 20.0 \\
    \midrule
    \midrule
    \multirow{2}*{Mute en} & en & 18.1 & 17.8 & 26.5 & 0.9 & 21.9 & 3.8 \\
    & zh & 0.0 & 38.6 & 10.8 & 1.3 & 4.9 & 4.6 \\
    \midrule
    \multirow{2}*{Mute zh} & en & 0.0 & 22.3 & 7.1 & 0.9 & 1.8 & 4.4 \\
    & zh & 63.7 & 18.4 & 102.8 & 19.0 & 81.1 & 12.8 \\
    \bottomrule
    \end{tabular}
    \caption{Language-based selective attack results on FLEURS test sets for English (en), French (fr), and Chinese (zh) under the $\epsilon=0.02$ constraint. For each group, we report WER or CER and its composition, as well as the percentage of muted samples ($\varnothing$) and average sequence length (\texttt{asl}).}
    \label{tab:lang_eplison}
\end{table}

Finally, the language-based selective attack results with the $\epsilon=0.02$ constraint are presented in Table~\ref{tab:lang_eplison}. Compared to the unbounded results in Table~\ref{tab:lang}, the effectiveness of the attack is largely reduced for audio from the targeted language, while the untargeted languages are generally less affected. This suggests that enforcing imperceptibility makes it more challenging to achieve strong selective suppression in this setting.


\end{document}